\documentclass[sigconf]{acmart}
\usepackage{graphicx}
\usepackage{subfigure}
\usepackage{multirow}
\usepackage{xcolor}

\AtBeginDocument{%
  \providecommand\BibTeX{{%
    \normalfont B\kern-0.5em{\scshape i\kern-0.25em b}\kern-0.8em\TeX}}}

\setcopyright{acmcopyright}

\copyrightyear{2022}
\acmYear{2022}
\setcopyright{acmlicensed}\acmConference[WSDM '22]{Proceedings of the Fifteenth ACM International Conference on Web Search and Data Mining}{February 21--25, 2022}{Tempe, AZ, USA}
\acmBooktitle{Proceedings of the Fifteenth ACM International Conference on Web Search and Data Mining (WSDM '22), February 21--25, 2022, Tempe, AZ, USA}
\acmPrice{15.00}
\acmDOI{10.1145/3488560.3498437}
\acmISBN{978-1-4503-9132-0/22/02}

\begin{document}
\fancyhead{}
\title{DualDE: Dually Distilling Knowledge Graph Embedding for Faster and Cheaper Reasoning}


\author{Yushan Zhu \textsuperscript{*}}
\affiliation{%
  \institution{Zhejiang University}
  \city{Hangzhou}
  \state{Zhejiang}
  \country{China}
}
\email{yushanzhu@zju.edu.cn}

\author{Wen Zhang \textsuperscript{*}}
\affiliation{%
  \institution{Zhejiang University}
  \city{Hangzhou}
  \state{Zhejiang}
  \country{China}
}
\email{wenzhang2015@zju.edu.cn}

\author{Mingyang Chen}
\affiliation{%
  \institution{Zhejiang University}
  \city{Hangzhou}
  \state{Zhejiang}
  \country{China}
}
\email{mingyangchen@zju.edu.cn}

\author{Hui Chen}
\affiliation{%
  \institution{Alibaba Group}
  \city{Hangzhou}
  \state{Zhejiang}
  \country{China}}
  \email{weidu.ch@alibaba-inc.com}

\author{Xu Cheng}
\affiliation{%
  \institution{Peking University}
  \city{Beijing}
  \country{China}}
  \email{chengxu@pku.edu.cn}

\author{Wei Zhang}
\affiliation{%
  \institution{Alibaba Group}
  \city{Hangzhou}
  \state{Zhejiang}
  \country{China}}
  \email{zhangweinus@gmail.com}
  
\author{Huajun Chen \textsuperscript{\textsection}}
\affiliation{%
  \institution{College of Computer Science \\ Hangzhou Innovation Center \\ Zhejiang University}
  \city{Hangzhou}
  \state{Zhejiang}
  \country{China}}
\email{huajunsir@zju.edu.cn}  

\renewcommand{\shortauthors}{Zhu, et al.}


\begin{abstract}
  Knowledge Graph Embedding (KGE) is a popular method for KG reasoning and training KGEs with higher dimension are usually preferred since they have better reasoning capability. However, high-dimensional KGEs pose huge challenges to storage and computing resources and are not suitable for resource-limited or time-constrained applications, for which faster and cheaper reasoning is necessary. To address this problem, we propose DualDE, a knowledge distillation method to build low-dimensional student KGE from pre-trained high-dimensional teacher KGE. DualDE considers the dual-influence between the teacher and the student. In DualDE, we propose a soft label evaluation mechanism to adaptively assign different soft label and hard label weights to different triples, and a two-stage distillation approach to improve the student's acceptance of the teacher. Our DualDE is general enough to be applied to various KGEs.
  Experimental results show that our method can successfully reduce the embedding parameters of a high-dimensional KGE by 7$\times$-15$\times$ and increase the inference speed by 2$\times$-6$\times$ while retaining a high performance. We also experimentally prove the effectiveness of our soft label evaluation mechanism and two-stage distillation approach via ablation study.
\end{abstract}




\maketitle
\begingroup\renewcommand\thefootnote{*}
\footnotetext{Equal contribution.}
\begingroup\renewcommand\thefootnote{\textsection}
\footnotetext{Corresponding author.\\http://zjukg.org}


\section{Introduction}
\noindent Knowledge Graph (KG) is composed of triples representing facts in the form of \emph{(head entity, relation, tail entity)}, abbreviate as \emph{(h, r, t)}. KGs have proved to be useful for various AI tasks, such as semantic  search~\cite{6_DBLP:conf/emnlp/BerantCFL13,7_DBLP:conf/acl/BerantL14}, information extraction~\cite{8_DBLP:conf/acl/HoffmannZLZW11,9_DBLP:conf/i-semantics/DaiberJHM13} and question answering~\cite{10_DBLP:journals/corr/ZhangLHJLW016,11_DBLP:conf/www/DiefenbachSM18}. However, it is well known that KGs are usually far from complete and this motivates many researches for KG completion, among which a common and widely used series  of  methods is Knowledge Graph Embedding (KGE), such as TransE \cite{14_DBLP:conf/nips/BordesUGWY13}, ComplEx \cite{12_DBLP:conf/icml/TrouillonWRGB16}, and RotatE \cite{19_DBLP:conf/iclr/SunDNT19}. To achieve better performance, as shown in Figure \ref{fig_changes}, training KGEs with higher dimension is typically preferred. 

But embeddings with lower dimensions provide obvious or even indispensable conveniences. The model size, i.e. the number of parameters, and the cost of reasoning time usually increase fast as the embedding dimension goes up. As shown in Figure \ref{fig_changes}, more and more little performance gain is got with larger embedding dimension, while the model size and reasoning cost keep increase linearly.
\begin{figure}[]
\setlength{\abovecaptionskip}{0.15cm}
\setlength{\belowcaptionskip}{-0.3cm} 
  \centering
  \small
    \subfigure[MRR and model size]{\includegraphics[width=0.236\textwidth]{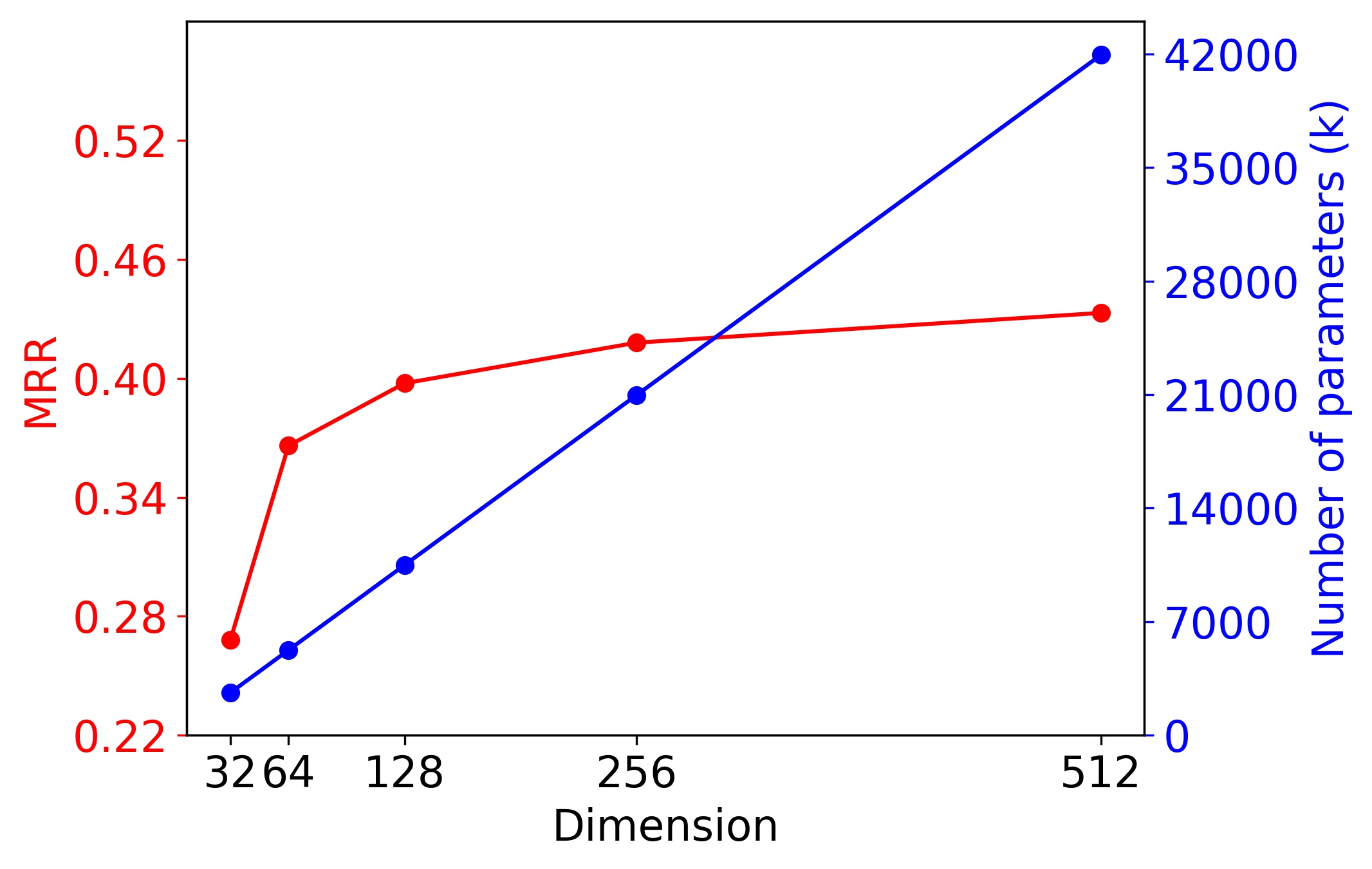}} 
    \subfigure[MRR and reasoning cost]{\includegraphics[width=0.236\textwidth]{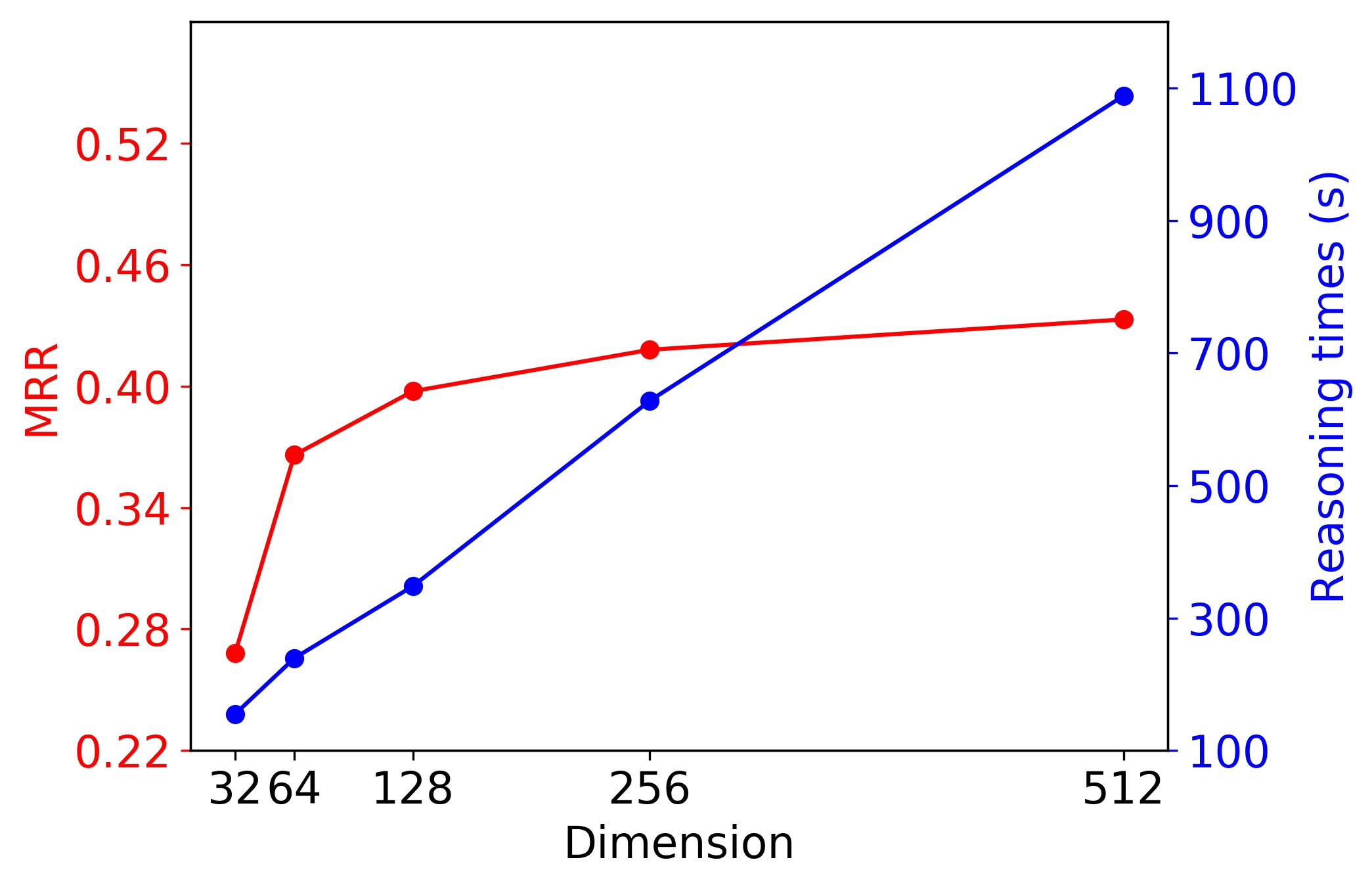}} \\
  \caption{The changes of performance (MRR), model size and reasoning cost  along the growth of embedding dimensions on WN18RR with ComplEx.} 
    \label{fig_changes}
\end{figure}
In addition, high-dimensional embeddings are impractical in many real-life scenarios. For example, a pre-trained billion-scale knowledge graph is expected to be fine-tuned to solve downstream tasks and deployed frequently at a cheaper cost. For applications with limited computing resources such as deploying KG on edge computing or mobile devices, or with limited time for reasoning such as online financial predictions, KG embedding with lower dimensions is indispensable.

However, directly training a model with a small embedding size normally performs poorly as shown in Figure~\ref{fig_changes}. 
We propose a new research question \textbf{: is it possible to learn low-dimensional KGEs from pre-trained high-dimensional ones so that we could achieve good performance as long as faster and cheaper inference?}

Knowledge Distillation~\cite{2_DBLP:journals/corr/HintonVD15} is a widely used technology to learn knowledge from a large model (teacher) to build a smaller model (student).
The student learns from both the ground-truth labels and the soft labels from the teacher. 
In this work, we propose a novel KGE distillation method
, named \textbf{DualDE}, which is capable of distilling essence from a high-dimensional KGE into a smaller embedding size with only a little or no loss of accuracy.
In DualDE, we dually distilling KGE considering the \textbf{dual-influence between the teacher and the student}: (1) the teacher's influence on the student and (2) the student's influence on the teacher.


For \textit{the teacher's influence on the student}, it is well known that the soft labels output by the teacher will affect the student. While in many previous distillation works~\cite{2_DBLP:journals/corr/HintonVD15,DBLP:conf/cvpr/ParkKLC19,DBLP:conf/aaai/MirzadehFLLMG20,DBLP:conf/www/Wang0MS21}, all samples have the same hard and soft label weight, they does not distinguish the quality of soft labels of different samples from the teacher. In fact, KGE methods have different levels of mastery of different triples~\cite{19_DBLP:conf/iclr/SunDNT19}. For some triples that are difficult to be mastered by KGE methods, they usually cannot obtain reliable scores~\cite{19_DBLP:conf/iclr/SunDNT19}. Making the student imitate the teacher with unreliable scores of these difficult triples will bring negative impacts on them. To obtain a better distillation effect, we propose that the student should be able to judge the quality of the soft labels provided by the teacher and selectively learn from them, rather than treating them equally. We introduce a \textbf{soft label evaluation mechanism} into DualDE to evaluate the quality of the soft labels provided by the teacher, and adaptively assigns different soft label and hard label weights to different triples, which will retain the positive effect of high-quality soft labels and avoid the negative impact of low-quality soft labels.

For \textit{the student's influence on the teacher}, it has not been study enough in previous works. Sun~\cite{5_DBLP:conf/emnlp/SunCGL19} proved that the overall performance also depends on the student's acceptance of the teacher. We hope to constantly adjust the teacher 
according to the student's current learning situation, so as to make the teacher more acceptable to the student and improve the final distillation result. Thus, we propose a \textbf{two-stage distillation approach} into DualDE to improve the student's acceptance of the teacher by adjusting the teacher according to the student's output. 
The basic idea is that although the pre-trained teacher is already strong,  it may not be the most suitable one for the current student. A teacher who has a similar output distribution with the student is more conducive to the student's learning~\cite{DBLP:conf/www/Wang0MS21}. 
Therefore, in addition to a standard distillation stage in which the teacher is always static, we devise a second stage distillation in which the teacher is unfrozen and tries to learn from its student in reverse to become more acceptable for the student.

We evaluate DualDE with several typical KGEs and standard 
KG datasets. Results prove the effectiveness of our method, showing that (1) the low-dimensional KGEs distilled by DualDE perform much better than the same sized KGEs  directly trained and only a little or not worse than original high-dimensional KGEs; 
(2) the low-dimensional KGEs distilled by DualDE infer significantly faster than original high-dimensional KGEs;
(3) our proposed soft label evaluation mechanism and two-stage distillation approach work well and further improve the distillation results.

In summary, our contributions are three-fold:
\begin{itemize}
\item We propose a novel framework to distill lower-dimensional KGEs from higher-dimensional ones and it achieves good performance.
\item We consider the dual-influence between the teacher and the student in the distillation process, and propose a soft label evaluation mechanism  to distinguish the quality of soft labels of different triples and a two-stage distillation to improve the student's adaptability to the teacher.
\item We experimentally prove that our proposal can reduce embedding parameters of a high-dimensional KGE by \textbf{7-15 times} and increase the inference speed about \textbf{2-6 times} with only a little or no performance loss.
\end{itemize}

\section{Related Work}

\subsection{Knowledge Graph Embedding}
In recent years, KGE technology has been rapidly developed and applied. Its key idea is to transform entities and relations of KGs into a continuous vector space as vector representations. And then the embeddings can be further applied to various KG downstream tasks. RESCAL \cite{18_DBLP:conf/icml/NickelTK11} is the first relation learning method based on tensor decomposition. 
To improve RESCAL, DistMult \cite{22_DBLP:journals/corr/YangYHGD14a} restricts the relation matrix to a diagonal matrix to simplify the model,
ComplEx \cite{12_DBLP:conf/icml/TrouillonWRGB16} embeds entities and relations into the complex space to model asymmetric relations, and SimplE \cite{DBLP:conf/nips/Kazemi018} solves the independence problem of embedding vectors in tensor decomposition.
TransE \cite{14_DBLP:conf/nips/BordesUGWY13} is the first translation-based KGE method and regards the relation as a translation from the head entity to the tail entity. And various variants of TransE have been proposed. 
TransH \cite{25_DBLP:conf/aaai/WangZFC14} proposes that an entity should have different representations with different relations. TransR \cite{26_DBLP:conf/aaai/LinLSLZ15} believes that different relations pay attention to different attributes of entities. TransD \cite{27_DBLP:conf/acl/JiHXL015} demonstrates that a relation may represent multiple semantics. 
In addition, rotation models such as RotatE \cite{19_DBLP:conf/iclr/SunDNT19}, QuatE \cite{20_DBLP:conf/nips/0007TYL19}, and DihEdral \cite{21_DBLP:conf/acl/XuL19} 
regard the relation as the rotation between the head and tail entity.

Although the KGEs are simple and effective, there is an obvious problem that high-dimensional KGEs pose a huge challenge to storage and computing. It is necessary to reduce the dimension of KGEs and still retain a good performance for many practical application scenarios. However, there are very few researches on KGE compression. 
\cite{35_DBLP:conf/acl/Sachan20} proposes a method based on quantization technology to reduce the size of KGEs by representing entities as vectors of discrete codes.
However, quantization cannot improve the inference speed and often increases the difficulty of model convergence \cite{DBLP:conf/iccv/GongLJLHLYY19}. 
MulDE~\cite{DBLP:conf/www/Wang0MS21} is the first work to apply knowledge distillation to KGE. This method transfers the knowledge from multiple teachers to a student, but it requires pre-training multiple teacher models with different KGEs.
In this work, we propose an effective KGE compression method based on knowledge distillation considering the dual-influence between the teacher and the student.

\vspace{-3pt} 
\subsection{Knowledge Distillation}
In the last few years, the acceleration and compression of models have attracted a lot of research works. Common methods include network pruning \cite{32_DBLP:journals/tnn/CastellanoFP97,
33_DBLP:conf/iclr/MolchanovTKAK17}, quantification \cite{34_DBLP:conf/icml/LinTA16,35_DBLP:conf/acl/Sachan20}, parameters sharing \cite{36_DBLP:conf/iclr/DehghaniGVUK19,
37_DBLP:conf/iclr/LanCGGSS20}, and knowledge distillation \cite{2_DBLP:journals/corr/HintonVD15}.

Among them, knowledge distillation (KD) has been widely used in Computer Vision and Natural Language Processing since it can effectively reduce the model size and increase the model's inference speed. 
Its core idea is to use the teacher's output to guide the training of the student. What's more, KD has an advantage different from the other model compression methods mentioned above: 
different kinds of distillation targets can be designed according to needs, providing more modeling freedom. 
\cite{39_DBLP:journals/corr/abs-1903-12136} proposes to distill the pre-trained language model BERT \cite{44_DBLP:conf/naacl/DevlinCLT19} into a single-layer bidirectional long and short-term memory network.
\cite{5_DBLP:conf/emnlp/SunCGL19} proposes to enable students to fit the middle layer output of the teacher, instead of just the softmax layer output. \cite{38_DBLP:conf/iclr/TianKI20} believes that there are dependencies between the dimensions of data representation, and proposes maximizing the mutual information of the data representation of the student and the teacher.
\cite{45_DBLP:journals/corr/abs-1909-11687} gives up the transfer of BERT's the softmax layer and directly approximates the corresponding weight matrix of the student and the teacher.
\cite{DBLP:conf/cvpr/ParkKLC19} focuses on extracting the differences between samples rather than the information of a single sample itself, and proposes distance-wise loss and angle-wise distillation loss.
\cite{DBLP:conf/aaai/MirzadehFLLMG20} thinks that too-large size difference between two models is harmful for the effect of distillation
, and suggests using a medium-scale one to bridge this gap.


However, current KD methods cannot model the  dual-influence between the teacher and the student. 
In DualDE, for the teacher's influence on the student, we design a soft label evaluation mechanism to distinguish the quality of soft labels of different triples, and for the student's influence on the teacher, we proposed a two-stage distillation to improve the student's adaptability to the teacher.

\section{Method}

In this section, we introduce our KGE distillation method DualDE, in which a larger size KGE is regarded as \textit{teacher} and a small size KGE as \textit{student}. DualDE follows the typical training mechanism of knowledge distillation~\cite{2_DBLP:journals/corr/HintonVD15,39_DBLP:journals/corr/abs-1903-12136,44_DBLP:conf/naacl/DevlinCLT19} that the student is firstly encouraged to fit the hard labels from data with a hard labels loss, and then imitate the teacher via fitting soft labels from the teacher with a soft label loss. 
In DualDE, we make the student imitate teacher from following two aspects: overall credibility and embedding structure of target triples, since they contain the primary information~\cite{DBLP:conf/cvpr/ParkKLC19} and captures an invariant property of a model~\cite{DBLP:journals/jmlr/AchilleS18}. 

Different from typical knowledge distillation methods, dual-influence between the student and the teacher is fully explored in DualDE which includes teacher's influence on the student and student's influence on the teacher.

For \textit{the teacher's influence on the student}, it is well known that soft labels output by the teacher will affect the student. While in many previous works~\cite{2_DBLP:journals/corr/HintonVD15,DBLP:conf/cvpr/ParkKLC19,DBLP:conf/aaai/MirzadehFLLMG20,DBLP:conf/www/Wang0MS21}, all samples have the same hard and soft label weight, and they do not distinguish the quality of soft labels of different samples from the teacher. In fact, for triples that are difficult to be mastered by KGE methods, they often cannot output reliable scores~\cite{19_DBLP:conf/iclr/SunDNT19}. 
Making the student imitate the teacher with unreliable scores of triples will bring negative impacts on the student. 
We propose that the student should be able to judge the quality of the soft labels provided by the teacher and selectively learn from them, rather than treating them equally. Thus we introduce a \textbf{soft label evaluation mechanism} into DualDE to evaluate the quality of soft labels which will retain the positive effect of high-quality soft labels and avoid the negative impact of low-quality soft labels.

For \textit{the student's influence on the teacher}, it has not been studied enough in previous works. MulDE~\cite{DBLP:conf/www/Wang0MS21} pointed that the student absorbs the knowledge better from the teacher having a more similar output distribution with the student, which supports that there are suitable teachers and unsuitable teachers for the student. To make the teacher a more suitable teacher, 
different from previous works keeping teacher fixed all the time, we propose a \textbf{two-stage distillation approach} into DualDE. Conventional training of the student with the teacher frozen is referred to as the first stage.
In the second stage, the teacher is unfrozen and adjusted according to the student's situation. 
The basic idea is that we not only train the teacher with a hard label to guarantee its performance, but also engage it to fit a soft label generated from the student. Essentially, this can be regarded as a process that the teacher learns from its student in reverse. As a result, the teacher will become more adaptable to the student, thereby improving the distillation effect. 

Overall, DualDE is trained based on following loss 
\begin{equation}
\vspace{-2pt} 
\begin{aligned}
&L= L_{Stu}+ \gamma L_{Tea},\\
L_{Stu} = L_{Hard}^S &+ L_{Soft}^S, \; L_{Tea} = L_{Hard}^T + L_{Soft}^T,\\
\end{aligned}
    \label{dis_loss_s2}
\end{equation}
where $\gamma=0$ in the first distillation stage, and $\gamma=1$ in the second distillation stage.


Next, we elaborate on DualDE. Firstly, we define the KGE Distillation Objective. We then introduce the soft label evaluation mechanism and the two-stage distillation approach in detail. The model framework of DualDE is shown in Figure \ref{fig:model}.

\begin{figure*}[]
\centering
\vspace{-2pt}  
\setlength{\belowcaptionskip}{-0.3cm} 
    \centering
    \includegraphics[width=0.8\textwidth]{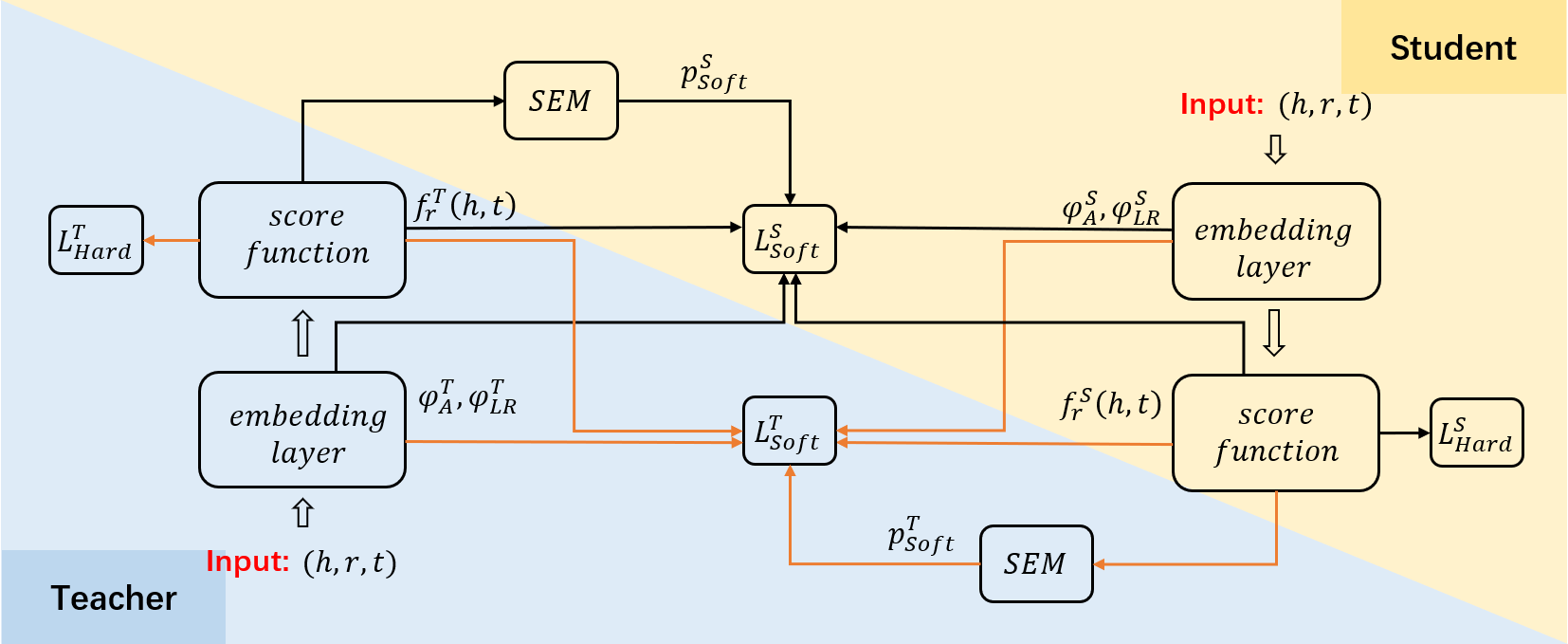}
    \caption{Model framework of DualDE. \textit{SEM} refers to the soft label evaluation module. The data stream with the black arrow ``\textcolor{black}{\textbf{$\to$}}'' participates in both the first and second stages of distillation, and the data stream with the orange arrow ``\textcolor{orange}{\textbf{$\to$}}'' only participates in the second stage of distillation.}
    \label{fig:model}
\end{figure*}

\subsection{KGE Distillation Objective}

Given a KG $\mathcal{K}= \{ E, R, T\} $, where $E$, $R$ and $T$ are the set of entities, relations and triples respectively. A KGE learns to express the relationships between entities in a continuous vector space. Specifically, for a triple $(h, r, t)$, where $h, t \in E$, $r \in R$, the KGE model could assign a score to it by a score function $f_r(h,t)$, to indicate the existence of $(h, r, t)$. Table \ref{t_score} summarizes the score function of some popular KGE methods.

\begin{table}[htbp]
\setlength{\abovecaptionskip}{0.2cm}
\caption{Score functions of some popular knowledge graph embedding methods. Here, $ <x^1,..., x^k> = \sum_i x^1_i...x_i^k $denotes the generalized dot product, $\bar{x}$ represents the conjugate of a complex number $x$, $\bullet$ represents the Hadamard product.}
\label{t_score}
\centering

\resizebox{0.41\textwidth}{!}{

\begin{tabular}{c | c }
\toprule
\textbf{Method}  & \textbf{Score function} $f_r(h,t)$ \\
\midrule
TransE  &$-\left \|h+r-t\right \|_p$   \\

SimplE  & $ (<h^{(H)}, r, t^{(T)}> + <t^{(H)}, r^{(inv)}, h^{(T)}>) /2$ \\ 

ComplEx &$Re(h^\top diag(r)\bar{t})$  \\

RotatE  &$-\left \| h\bullet r-t \right \|^2$   \\
\bottomrule
\end{tabular}
}

\end{table}

\subsubsection{Hard Label Loss.}
The hard label loss for the student is the original loss of the KGE method, usually a binary cross entropy loss:
\begin{equation}
\label{L_hard_loss}
\begin{aligned}
L_{Hard}^S=& -\sum_{(h,r,t)\in T\cup T^-} (y\log \sigma ( f_r^S{(h,t)}) \\
& + (1-y)(1-\log \sigma(f_r^S{(h,t)}))),\\
\end{aligned}
\end{equation}
where $f_r^S(h,t)$ is the score for triple  $(h, r, t)$ given by the student. $\sigma$ is the Softmax function. $y$ is the ground-truth label of $(h,r,t)$, and it is $1$ for positive triple $(h, r, t)$ and $0$ for negative triple $(h^\prime, r, t^\prime)$. $(h^\prime, r, t^\prime)$ is generated by randomly replacing $h$ or $t$ in $(h,r,t)\in T$ with $h'$ or $t'$, which could be expressed as

\begin{equation}
\setlength{\abovedisplayskip}{0pt}
\setlength{\belowdisplayskip}{10pt}
\label{sample_neg}
\begin{aligned}
T^-=& \{(h',r,t) \notin T|h' \in E \wedge h'\ne h\} \\
& \cup \{(h,r,t') \notin T|t' \in E \wedge t'\ne t \}.
\end{aligned}
\end{equation}

\subsubsection{Soft Label Loss}
In DualDE, we enable the student to learn two kinds of knowledge from the teacher: the credibility and the embedding structure of triples.

The credibility of a triple is reflected by its score output by the KGE model, and the score difference between the student and the teacher is defined as
\begin{equation}
\label{L_score}
    d_{Score}=l_{\delta}(f_{r}^{T}(h,t),f_{r}^{S}(h,t)).
\end{equation}
The embedding structure of a triple can be reflected by the length ratio and the angle between the embedding vectors of the head entity $h$ and tail entity $t$~\cite{DBLP:conf/cvpr/ParkKLC19}. The embedding structure difference between the teacher and the student is defined as  
\begin{equation}
\label{loss_struc}
\begin{aligned}
    d_{Structure}=& l_{\delta}(\varphi _{A}(h^{T},t^{T}),\varphi _{A}(h^{S},t^{S}))  \\& + l_{\delta}(\varphi _{LR}(h^{T},t^{T}),\varphi _{LR}(h^{S},t^{S})),
\end{aligned}
\end{equation}
where $f_r^T(h,t)$ ($f_r^S(h,t)$) is the score for triple $(h, r, t)$ given by the teacher (student), $\varphi _{A}(h,t)=\langle \frac{h}{\left \| h \right \|_2},\frac{t}{\left \| t \right \|_2} \rangle$ and $\varphi _{LR}(h,t)=\frac{\left \| h \right \|_2}{\left \| t \right \|_2}$, $l_{\delta}$ is Huber loss with $\delta=1$, $h^T(t^T)$ and $h^S(t^S)$ is the head (tail) entity embedding of the teacher and the student respectively, $l_{\delta}$ is Huber loss with $\delta=1$, which is defined as
\begin{equation}
    \label{huber_loss}
    l_{\delta}(a,b)=\left\{\begin{matrix}
\frac{1}{2}(a-b)^{2}, & \left | a-b \right |\leqslant 1,\\ 
 \left | a-b \right |-\frac{1}{2},& \left | a-b \right |>  1.
\end{matrix}\right.
\end{equation}


We combined the triple score difference and embedding structure difference between the student and the teacher as the soft label optimization goal:
\begin{equation}
    d_{Soft}=d_{Score}+d_{Structure}.
\end{equation}

\subsection{Soft Label Evaluation Mechanism}
We propose the soft label evaluation mechanism to evaluate the quality of the soft labels provided by the teacher, and adaptively assign different soft label and hard label weights to different triples, so as to retain the positive effect of high-quality soft labels and avoid the negative impact of low-quality soft labels.

Theoretically, the KGE model will give higher scores to positive triples and lower scores to negative triples, but it is opposite for some triples that are difficult to be mastered by the KGE model. 
Specifically, if the teacher gives a high (low) score to a negative (positive) triple, which means the teacher tends to judge it as a positive (negative) triple, the soft label of this triple output by the teacher is unreliable and misleading and may have a negative impact on the student. For this triple, we need to weaken the weight of the soft label and encourage the student to learn more from the hard label.

The soft label weights of the student for positive triples and negative triples are defined as Eq. (\ref{pos_soft_weight}) and Eq. (\ref{neg_soft_weight}), respectively:
\begin{equation}
    p^S_{PosSoft}=\frac{1}{1+e^{-\alpha_1 (f_r^T(h,t)+\beta_1)}}
    \label{pos_soft_weight},
\end{equation}
\begin{equation}
    p^S_{NegSoft}=1-\frac{1}{1+e^{-\alpha_2 (f_r^T(h,t)+\beta_2)}}
    \label{neg_soft_weight},
\end{equation}
where $\alpha_1$, $\beta_1$, $\alpha_2$ and $\beta_2$ are learned from training data. The student's final soft label loss and hard label loss can be expressed as Eq. (\ref{dis_loss_stu_soft}) and Eq. (\ref{dis_loss_stu_hard}), respectively:
\begin{equation}
\begin{aligned}
L^S_{Soft}=&\sum_{(h,r,t)\in T} p^S_{PosSoft}\cdot  d_{soft} + \sum_{(h,r,t)\in T^-} p^S_{NegSoft}\cdot  d_{soft},
\end{aligned}
    \label{dis_loss_stu_soft}
\end{equation}

\begin{equation}
\begin{aligned}
L^S_{Hard}=
& \sum_{(h,r,t)\in T} (1-p^S_{PosSoft}) \cdot \log \sigma ( f_r^S{(h,t)}) \\
& + \sum_{(h,r,t)\in T^-} (1-p^S_{NegSoft}) \cdot (1-\log \sigma(f_r^S{(h,t)})).
\end{aligned}
    \label{dis_loss_stu_hard}
\end{equation}

By evaluating the quality of the teacher's score for each triple, different soft label weights and hard label weights are given to different triples adaptively, helping the student selectively learn the knowledge from the teacher and get better performance. In addition, this method can balance the soft label loss and the hard label loss automatically without defining any hyperparameter manually.

\subsection{Two-stage Distillation Approach}
In the previous part, we introduced how to enable the student to extract knowledge from the KGE teacher, where the student is trained with hard labels and the soft labels generated by a fixed teacher. To obtain a better student, we propose a two-stage distillation approach to improve the student's acceptance of the teacher by unfreezing the teacher and engage it to learn from the student in the second stage of distillation. 

\subsubsection{The First Stage.} The first stage is similar to conventional knowledge distillation methods in which the teacher is frozen and unchanged when training the student. 
The final loss of the first stage is Eq. (\ref{dis_loss_s2}) with $\gamma=0$.

\subsubsection{The Second Stage.} 

While adjusting the teacher in this stage, for the triples that the student does not mastered well, we also hope to reduce the negative impact of the output of the student on the teacher, and make the teacher learn more from hard labels, so as to maintain the teacher's high accuracy. Thus, we also apply the soft label evaluation mechanism in the adjustment of teacher. By evaluating the score given by the student to each triple, the weights of hard labels and soft labels for the teacher are allocated adaptively.

Similarly, the soft label weights of the teacher for positive triples and negative triples are defined as Eq. (\ref{pos_soft_weight_tea}) and Eq. (\ref{neg_soft_weight_tea}), respectively:

\begin{equation}
    p^T_{PosSoft}=\frac{1}{1+e^{-\alpha_3 (f_r^S(h,t)+\beta_3)}},
    \label{pos_soft_weight_tea}
\end{equation}

\begin{equation}
    p^T_{NegSoft}=1-\frac{1}{1+e^{-\alpha_4 (f_r^S(h,t)+\beta_4)}},
    \label{neg_soft_weight_tea}
\end{equation}
where $\alpha_3$, $\beta_3$, $\alpha_4$ and $\beta_4$ are learned from training data. The teacher's final soft label loss and hard label loss can be expressed as Eq. (\ref{dis_loss_tea_soft}) and Eq. (\ref{dis_loss_tea_hard}), respectively:
\begin{equation}
\begin{aligned}
L^T_{Soft}=&\sum_{(h,r,t)\in T} p^T_{PosSoft}\cdot  d_{soft} + \sum_{(h,r,t)\in T^-} p^T_{NegSoft}\cdot  d_{soft},
\end{aligned}
    \label{dis_loss_tea_soft}
\end{equation}

\begin{equation}
\begin{aligned}
L^T_{Hard}=
&\sum_{(h,r,t)\in T} (1-p^T_{PosSoft}) \cdot \log \sigma ( f_r^T{(h,t)}) \\
& + \sum_{(h,r,t)\in T^-} (1-p^T_{NegSoft}) \cdot (1-\log \sigma(f_r^T{(h,t)})).
\end{aligned}
    \label{dis_loss_tea_hard}
\end{equation}

The final loss of the second stage is Eq. (\ref{dis_loss_s2}) with $\gamma=1$.

\section{Experiments}
We evaluate DualDE on typical KGE benchmarks, and are particularly interested in the following questions:
\begin{itemize}
    \item Is DualDE capable of distilling a good low-dimensional student from the high-dimensional teacher and performing better than the same dimensional model trained from scratch without distillation or using other KD methods?
    \item How much is the inference time improved after distillation?
    \item Do the soft label evaluation mechanism and two-stage distillation approach contribute to our proposal and how much?
\end{itemize}

\subsection{Datasets and Implementation Details}
\subsubsection{Datasets.}
We experiment on two common knowledge graph completion benchmark datasets WN18RR \cite{16_DBLP:conf/emnlp/ToutanovaCPPCG15} and FB15k-237 \cite{15_DBLP:conf/aaai/DettmersMS018}, subsets of WordNet \cite{14_DBLP:conf/nips/BordesUGWY13} and Freebase  \cite{14_DBLP:conf/nips/BordesUGWY13} with redundant inverse relations eliminated. Table~\ref{table_dataset} shows the statistics of these two datasets.

\begin{table}[!hbpt]

\caption{Statistics of datasets we used in the experiments.}\label{table_dataset}

\resizebox{0.42\textwidth}{!}{
\begin{tabular}{llllll}
\toprule
\textbf{Dataset}   & \textbf{\#Ent.} & \textbf{\#Rel.} & \textbf{\#Train} & \textbf{\#Valid} & \textbf{\#Test} \\
\midrule
WN18RR    & 40,943      & 11         & 86,835  & 3,034   & 3,134  \\
FB15k-237 & 14,541      & 237        & 272,115 & 17,535  & 20,466 \\
\bottomrule
\end{tabular}
}
\end{table}

\subsubsection{Evaluation Metrics.}
We adopt standard metrics MRR, and Hit@$k$ $(k=1,3,10)$. Given a test triple $(h, r, t)$, we first replace the head entity $h$ with each entity $e \in E$ and generate candidate triples $(e, r, t)$. Then we use the score function $f_r(e,t)$ to calculate the scores of all candidate triples and arrange them in descending order, according to which, we obtain the rank of $(h,r,t)$, $rank_{h}$ as its head prediction result. For $(h,r,t)$'s tail prediction, we replace $t$ with all $e \in E$ to generate candidate triples $(h, r, e)$, and get the tail prediction rank $rank_{t}$ in a similar way. We average $rank_{h}$ and $rank_{t}$ as the final rank of $(h, r, t)$. Finally, we calculate MRR, and Hit@$k$ via the rank of all test triples. MRR is their mean reciprocal rank. And Hit@$k$ measures the percentage of test triples with rank $\le k$. We use the filtered setting \cite{14_DBLP:conf/nips/BordesUGWY13} by removing all triples in the candidate set that existing in training, validating, and testing sets. 

\begin{table*}[htbp]
\caption{Link prediction results on WN18RR. Bold numbers are the best results between different methods.}
\label{T-WN18RR}
\centering
\resizebox{0.9\textwidth}{!}{

\begin{tabular}{cc|cccc|cccc|cccc|cccc}
\toprule
\multirow{2}{*}{\textbf{Dim}} 
& \multirow{2}{*}{\textbf{Method}}
& \multicolumn{4}{c}{\textbf{TransE}}     & \multicolumn{4}{c}{\textbf{SimplE}}      & \multicolumn{4}{c}{\textbf{ComplEx}}  & 
\multicolumn{4}{c}{\textbf{RotatE}}  \\
 & & \multicolumn{1}{|c}{\emph{MRR}} & \emph{H10} & \emph{H3} & \emph{H1} & \emph{MRR}   & \emph{H10} & \emph{H3} & \emph{H1} & \emph{MRR}   & \emph{H10} & \emph{H3} & \emph{H1}  & \emph{MRR} & \emph{H10} & \emph{H3} & \emph{H1}\\
\midrule
512                         & Tea.        
& .232 & .531 & .410 & .025
& .421 & .485 & .433 & .389
& .433 & .515 & .458 & .387
& .477 & .575 & .498 & .427
\\
\hline
\multirow{5}{*}{64}        
& no-DS       
& .192 & .476& .325& .022
& .357 & .463& .397& .293
& .366 & .469& .408& .303
& .459 & .542& .475& .412

\\
& BKD 
&.214 &.498	&.365 &.023
&.378 &.459 &.408 &.328
&.377 &.475 &.429 &.330
&.462 &.551 &.476 &.417

\\

& RKD  
& .227 & .524 &.404& \textbf{.033} 
& .406 & .476 &.421 &.370
& .396 & .498 &.438 &.343
& .469 & .564 &.482 &.424
\\ 

& TA
& .224 & .517 &.384& .021
& .392 & .484 &\textbf{.434} &.327
& .407 & .494 &.439 &.362
& .467 & .552 &.481	&.422 \\

& DualDE
& \textbf{.230}	&\textbf{.528} &\textbf{.409} &.030
& \textbf{.412} &\textbf{.485} &.431 & \textbf{.366}
& \textbf{.419} &\textbf{.507} &\textbf{.445} &\textbf{.379}
& \textbf{.472} &\textbf{.568} &\textbf{.488} &\textbf{.427}
\\ 

\hline
\multirow{5}{*}{32}        
& no-DS       
&.164 &.410 &.259 &.023
&.273 &.396 &.286 &.186
&.268 &.366 &.296 &.216
&.421 &.453 &.441 &.401
\\

& BKD 
&.184 &.442 &.302 &.026
&.321 &.452 &.379 &.259
&.343 &.406 &.377 &.302
&.441 &.497 &.451 &.412
\\

& RKD  
&.194 &.454 &.325 &.028
&.372 &.475	&.411 &.297
&.368 &.456 &.397 &.322
&.455 &.529 &.470 &.416

\\

& TA
&.189 &.458 &.318 &.032
&.359 &.476 &.407 &.283
&.372 &.464 &.406 &.315
&.452 &.519 &.468 &.417

\\

& DualDE
&\textbf{.210} &\textbf{.484} &\textbf{.349} &\textbf{.035}
&\textbf{.384} &\textbf{.479} &\textbf{.423} &\textbf{.311}
&\textbf{.397} &\textbf{.473} &\textbf{.422} &\textbf{.352}
&\textbf{.468} &\textbf{.560} &\textbf{.486} &\textbf{.419}

\\

\bottomrule
\end{tabular}

}

\end{table*}
\begin{table*}[htbp]
\caption{Link prediction results on FB15k-237. Bold numbers are the best results between different methods.}
\label{T-FB15k-237}
\centering
\resizebox{0.9\textwidth}{!}{
\begin{tabular}{cc|cccc|cccc|cccc|cccc}
\toprule
\multirow{2}{*}{\textbf{Dim}} 
& \multirow{2}{*}{\textbf{Method}}
& \multicolumn{4}{c}{\textbf{TransE}}     & \multicolumn{4}{c}{\textbf{SimplE}}      & \multicolumn{4}{c}{\textbf{ComplEx}}  & 
\multicolumn{4}{c}{\textbf{RotatE}}  \\
 & & \multicolumn{1}{|c}{\emph{MRR}} & \emph{H10} & \emph{H3} & \emph{H1} & \emph{MRR}   & \emph{H10} & \emph{H3} & \emph{H1} & \emph{MRR}   & \emph{H10} & \emph{H3} & \emph{H1}  & \emph{MRR} & \emph{H10} & \emph{H3} & \emph{H1}\\
\midrule
512                         & Tea.        
& .286 &.481 &.328 &.185
& .283 &.454 &.309 &.199
& .298 &.472 &.327 &.213
& .326 &.520 &.361 &.229

\\
\hline
\multirow{5}{*}{64}        
& no-DS       
& .234 &.401 &.259 &.151
& .214 &.376 &.239 &.133
& .204 &.378 &.226 &.119
& .310 &.471 &.325 &.212

\\
& BKD 
& .250 &.415 &.277 &.168
& .244 &.407 &.273 &.160
& .258 &.422 &.293 &.187
& .307 &.478 &.333 &.216

\\

& RKD  
& .276 &.452 &.308 &.187
& .262 &.427 &.288 &.172
& .295 &.470 &.326 &.212
& .314 &.505 &.347 &.221
\\

& TA
& .271 &.440 &.299 &.176
& .254 &.421 &.282 &.167
& .282 &.464 &.322 &.205
& .309 &.489 &.339 &.215

\\


& DualDE
& \textbf{.279} &\textbf{.455} &\textbf{.312} &\textbf{.190}
& \textbf{.271} &\textbf{.431} &\textbf{.289} &\textbf{.174}
& \textbf{.303} &\textbf{.478} &\textbf{.334} &\textbf{.218}
& \textbf{.319}	&\textbf{.513} &\textbf{.358} &\textbf{.226}

\\
\hline
\multirow{5}{*}{32}        
& no-DS       
&.183 &.317 &.199 &.115
&.155 &.314 &.181 &.075
&.162 &.317 &.171 &.096
&.285 &.461 &.316 &.195

\\

& BKD 
&.223 &.378 &.243 &.146
&.187 &.331 &.205 &.112
&.239 &.394 &.262 &.161
&.294 &.457 &.312 &.214

\\

& RKD  
&.246 &.410 &.272 &.162
&.213 &.367 &.226 &.125
&.270 &.440 &.298 &.185
&.303 &.486 &.337 &.210

\\

& TA
&.242 &.405 &.267 &.158
&.208 &.362 &.213 &.119
&.259 &.423 &.281 &.177
&.296 &.475 &.327 &.206

\\

& DualDE
&\textbf{.254} &\textbf{.418} &\textbf{.280} &\textbf{.173}
&\textbf{.236} &\textbf{.407} &\textbf{.252} &\textbf{.149}
&\textbf{.274} &\textbf{.444} &\textbf{.301} &\textbf{.189}
&\textbf{.306} &\textbf{.489} &\textbf{.338} &\textbf{.216}

\\

\bottomrule
\end{tabular}
}

\end{table*}

\subsubsection{Baselines.}
We implement DualDE by employing four commonly used KGE models, including TransE \cite{14_DBLP:conf/nips/BordesUGWY13}, ComplEx \cite{12_DBLP:conf/icml/TrouillonWRGB16}, SimplE \cite{DBLP:conf/nips/Kazemi018} and RotatE \cite{19_DBLP:conf/iclr/SunDNT19}.

In addition to the directly trained student of required dimension without distillation (no-DS), 
\begin{itemize}
    \item BKD \cite{2_DBLP:journals/corr/HintonVD15} is the most basic and commonly used KD method. We use BKD by minimizing the KL divergence of the triple score distributions output by the teacher and student. 
    \item RKD \cite{DBLP:conf/cvpr/ParkKLC19} is a typical embedding-based approach, focusing on the structural differences between samples. 
    In the experiment, we jointly use the distance-wise and angle-wise distillation losses proposed in the original paper.
    \item TA~\cite{DBLP:conf/aaai/MirzadehFLLMG20} proposes a medium-scale network (Teaching Assistant) to bridge the gap between the two models. We choose the best TA size recommended by the authors, whose MRR is closest to the average MRR of the teacher and the student.
    \item MulDE~\cite{DBLP:conf/www/Wang0MS21} is the first work to apply KD technology to KGE, which proposes to transfer the knowledge from multiple teachers to a student. Since there is a big framework gap between MulDE which is based on multiple teachers and DualDE which is based on a single teacher, it is difficult to directly apply MulDE to the above 4 KGE methods and compare it with DualDE. To compare with MulDE fairly and reasonably, we modified DualDE to a multi-teacher framework similar to MulDE, called M-DualDE. Specifically, M-DualDE retains the same structure of four 64-dimensional teachers and one 32-dimensional student as MulDE, and uses the same KGE models as in MulDE. The difference is that M-DualDE replaces the three distillation strategies proposed in MulDE with our soft label evaluation mechanism and two-stage distillation approach, and finally calculates the weighted average of the soft labels from four teachers as the final soft label of the student according to the conventional method in multi-teacher distillation~\cite{DBLP:conf/icassp/WuCW19}.
\end{itemize}

The other experimental details of the baselines including hyperparameter settings are the same as their original papers.


\subsubsection{Implementation Details.}
We implement DualDE by extending OpenKE \cite{DBLP:conf/emnlp/HanCLLLSL18}, an open-source KGE framework based on PyTorch. We set embedding dimension $d_{teacher}= \{256, 512, 1024\}$, $d_{student}=\{128, 64, 32, 16\}$, and make $d_{teacher} = 512$, $d_{student}=\{64, 32\}$ for primary experiment. We set batch size to $1024$ and maximum training epoch to $3000$. For each positive triple, we generate $64$ negative ones for WN18RR and $25$ for FB15k-237 in each training epoch. We choose Adam~\cite{49_DBLP:journals/corr/KingmaB14} as the optimizer, and learning rate decay and trigger decay threshold is set to $0.96$ and $10$. We perform a search on the initial learning rate in $\{0.0001,0.0005,0.001,0.01\}$ and report the results from the best one.

\subsection{Q1: Does our method successfully distill a good student? }

To verify whether DualDE successfully distills a good student, we first train a student with only hard label loss, marked as `no-DS', which is the same as training a same dimensional original KGE model. We also train same dimensional students using DualDE and other KD methods. We compare them on link prediction. Table \ref{T-WN18RR} and \ref{T-FB15k-237} shows the results on WN18RR and FB15k-237 of 32-dimensional  and 64-dimensional students with 512-dimensional teachers.

\subsubsection{Results Analysis.}
First we analyze the results on WN18RR in Table~\ref{T-WN18RR}. Table~\ref{T-WN18RR} shows that the performance of `no-DS' model decreases significantly as the embedding dimension reducing. For SimplE, compared with the 512-dimensional teacher, a 32-dimensional `no-DS' model only achieves $64.8\%$, $66.1\%$, and $47.8\%$ results on MRR, Hit@3, and Hit@1. 
And for ComplEx, the MRR decreases from $0.433$ to $0.268$ ($38.1\%$). This illustrates that directly training low dimensional KGEs produces poor results.

Compared with `no-DS', DualDE greatly improves the performance of 32-dimensional students. The MRR of TransE, SimplE, ComplEx and RotatE on WN18RR improves from $0.164$ to $0.21$ ($28.0\%$), from $0.273$ to $0.384$ ($40.7\%$), from $0.268$ to $0.397$ ($48.1\%$), and from $0.421$ to $0.468$ ($11.2\%$). On the basis of `no-DS', our 32-dimensional students achieve an \textbf{average improvement of} $\textbf{32.0\%}$, $\textbf{23.0\%}$, $\textbf{33.9\%}$, and $\textbf{46.7\%}$ on MRR, Hit@10, Hit@3, and Hit@1 among these four KGEs, finally reaching an \textbf{average level of} 	$\textbf{92.9\%}$, $\textbf{94.8\%}$, $\textbf{93.1\%}$, and $\textbf{102.3\%}$ of teacher's results on MRR, Hit@10, Hit@3, and Hit@1. 
We can also observe a similar result on FB15k-237 in  Table~\ref{T-FB15k-237}. The results show that DualDE can achieve \textbf{16 times} (512:32) embedding compression rate (CR) while retaining most of the performance of the teacher (more than $90\%$), in spite of some performance loss, which is still much better than training a low-dimensional model directly without any distillation.

More importantly, DualDE helps 64-dimensional students achieve almost the same good performance as the 512-dimensional teachers. Take WN18RR for instance, our 64-dimensional student with RotatE achieves $99.0\%$, $98.8\%$, $98.0\%$, and $100.0\%$ results of the teacher on MRR, Hit@10, Hit@3, and Hit@1. And among these four KGEs, our 64-dimensional students achieve an \textbf{average level of} $\textbf{98.2\%}$, $\textbf{99.2\%}$, $\textbf{98.6\%}$, and $\textbf{103.0\%}$ of teacher's results on MRR, Hit@10, Hit@3, and Hit@1.
A similar phenomenon could be found on FB15k-237 in Table~\ref{T-FB15k-237}, and particularly for ComplEx, the MRR, Hit@10, Hit@3 and Hit@1 of DualDE ($0.303$, $0.478$, $0.334$ and $0.218$) even surpass the teacher's ($0.298$, $0.472$, $0.327$ and $0.213$). The results show that DualDE can achieve \textbf{8 times} (512:64) embedding CR with very little or even no performance loss.

In addition, compared with different KD methods including BKD, RKD, TA (Table~\ref{T-WN18RR} and~\ref{T-FB15k-237}), and MulDE (Table~\ref{T-MulDE}), DualDE achieves the best performance in almost all settings. 

\begin{table}[htbp]
\caption{Link prediction results compared with MulDE \protect\cite{DBLP:conf/www/Wang0MS21}.}\label{T-MulDE}
\resizebox{.44\textwidth}{!}{
\begin{tabular}{c|ccc|ccc}
\toprule
\multicolumn{1}{c|}{\multirow{2}{*}{Method}} & \multicolumn{3}{c}{WN18RR}       & \multicolumn{3}{c}{FB15k-237}     \\
\multicolumn{1}{c|}{}                                 & \textit{MRR} & \textit{H10} & \textit{H1} & \textit{MRR} & \textit{H10} & \textit{H1} \\
\midrule
MulDE-TransH                                         & .267         & .540         & .094        & .328         & .511         & .236        \\
M-DualDE-TransH                                             & .259            & \textbf{.545}            & .049           & .324            & \textbf{.515}            & \textbf{.241}     \\
\hline
MulDE-DistH                                          & .460         & .545         & .417        & .326         & .509         & .235        \\
M-DualDE-DistH                                             & \textbf{.462}            & \textbf{.548}            & .408           & \textbf{.328}            & \textbf{.513}            & \textbf{.237}     \\
\hline
MulDE-RefH                                           & .479         & .569         & .434        & .325         & .508         & .233        \\
M-DualDE-RefH                                             & .476            & \textbf{.571}            & \textbf{.437}           & \textbf{.326}            & \textbf{.513}            & .227     \\
\hline
MulDE-RotH                                           & .481         & .574         & .433        & .328         & .515         & .237        \\
M-DualDE-RotH                                             & \textbf{.483}            & .572            & \textbf{.437}           & .328            & \textbf{.518}            & \textbf{.243}     \\
\bottomrule
\end{tabular}
}
\end{table}

\subsubsection{More Different Dimensional Teachers and Students.}

To further evaluate our DualDE with more different dimensions, we also conduct experiments on 256-dimensional  and 1024-dimensional teachers and 16-dimensional  and 128-dimensional students. Figure \ref{HeatMap_transe_mrr} shows a heatmap of MRR results with TransE on WN18RR.

It shows that (1) for 128-dimensional students, the higher dimensional teacher achieve slightly better results; (2) for 64-dimensional students, the higher-dimensional teacher does not necessarily achieve better results; and (3) for 32-dimensional  and 16-dimensional students, the higher-dimensional teacher achieves worse results. 
This indicates that our method's best compression capability is about 8 times. An intuition is that although a bigger teacher is more expressive, an overly high compression ratio may prevent the teacher from transferring important knowledge to the student. This analysis reveals that for an application where an especially low-dimensional student is required and suppose the required dimension is $d$, instead of choosing a very high-dimensional teacher with fantastic performance,  \textbf{it is better to choose a teacher with dimension $\le 8\times d$}, which helps obtain a better student and save more pretraining costs.

\begin{figure}[!hbpt]
\centering
\setlength{\belowcaptionskip}{-0.2cm} 
    \centering
    \includegraphics[width=0.3\textwidth]{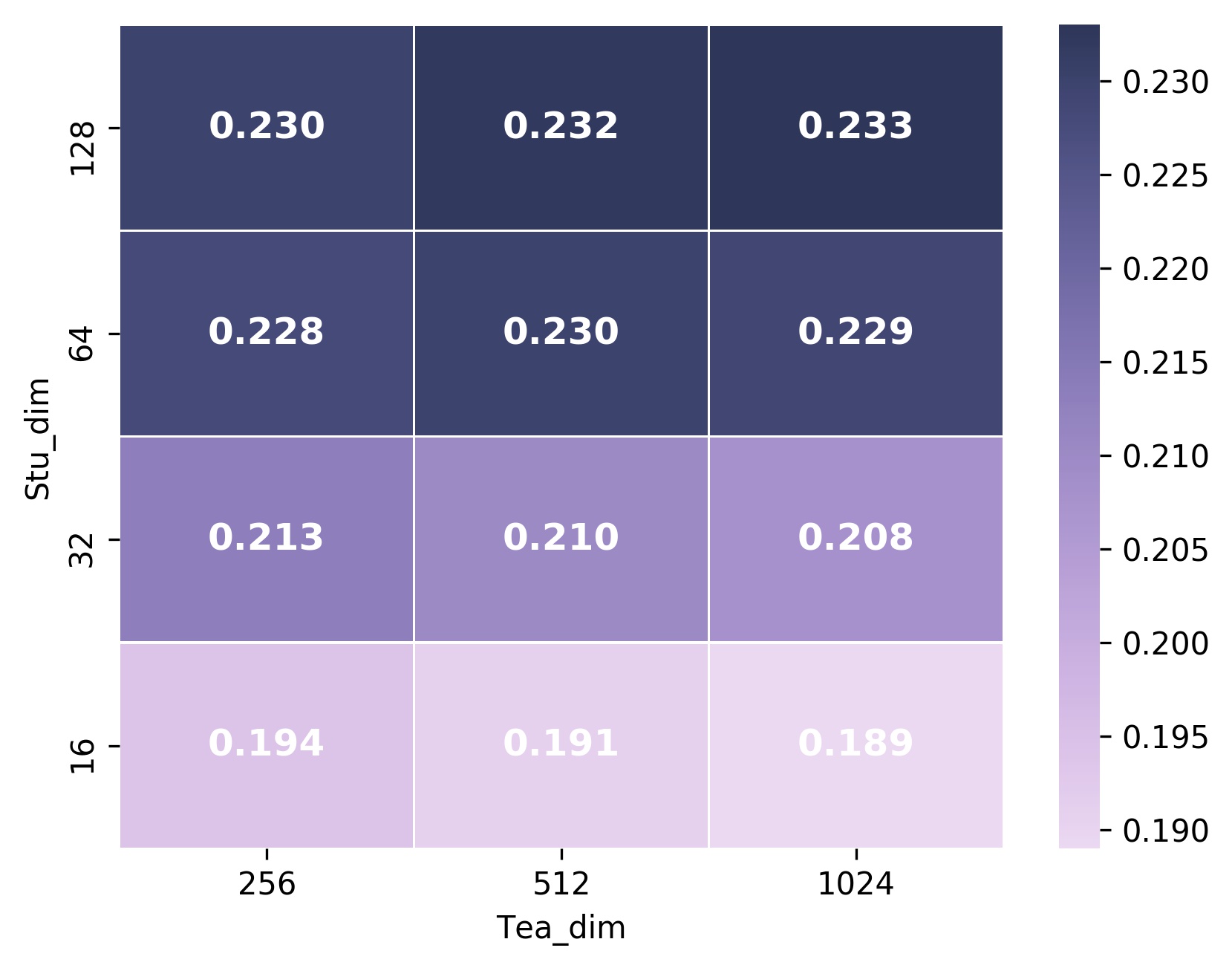}
    \caption{Students' test MRR distilled by teachers with different dimensions on the WN18RR with TransE.}
    \label{HeatMap_transe_mrr}
    \vspace{-0.05in}
\end{figure}

\subsection{Q2: Does the distilled student accelerate inference speed and how much?}

To test the inference speed, we conduct link prediction experiments on 93,003 samples from WN18RR and 310,116 samples from FB15k-237. Since the inference speed is not affected by the prediction mode (head or tail prediction), we uniformly compare the tail prediction time. The inference is performed on a single Tesla-V100 GPU, and the test batch size is set to the total number of entities: 40,943 for WN18RR and 14,541 for FB15k-237. To avoid accidental factors, we repeat the experiment 3 times and report the average time. 
Table~\ref{table_lp_efficiency} shows the result of inference time cost (in units of seconds).

\begin{table}[htbp]
\caption{The inference times (sec.).
}\label{table_lp_efficiency}
\resizebox{.43\textwidth}{!}{
\begin{tabular}{lc|cc|cc}
\toprule
\multicolumn{1}{l}{} & {Dim} & \multicolumn{2}{c|}{WN18RR} & \multicolumn{2}{c}{FB15k-237} \\
\midrule
\multirow{3}{*}{TransE}  & 512 (Tea.) & 417.5         & (1$\times$)          & 650.1          & (1$\times$)            \\
                         & 64  & 128.4         & (3.25$\times$)       & 199.8          & (3.25$\times$)         \\
                         & 32  & 93.5          & (4.47$\times$)       & 172.8          & (3.76$\times$)         \\
                         \hline
\multirow{3}{*}{SimplE}  & 512 (Tea.) & 530.2         & (1$\times$)          & 693.8          & (1$\times$)            \\
                         & 64  & 155.1           & (3.42$\times$)       & 230.4          & (3.01$\times$)         \\
                         & 32  & 107.8           & (4.92$\times$)       & 183.1          & (3.79$\times$)         \\
                         \hline
\multirow{3}{*}{ComplEx} & 512 (Tea.) & 1088.5        & (1$\times$)          & 1606.6         & (1$\times$)            \\
                         & 64  & 239.3         & (4.55$\times$)       & 337.2          & (4.76$\times$)         \\
                         & 32  & 154.9         & (7.03$\times$)       & 237.4          & (6.77$\times$)         \\
                         \hline
\multirow{3}{*}{RotatE}  & 512 (Tea.) & 1259.7        & (1$\times$)          & 1565.6         & (1$\times$)            \\
                         & 64  & 234.6         & (5.37$\times$)       & 341.2          & (4.59$\times$)         \\
                         & 32  & 161.3         & (7.81$\times$)       & 273.9            & (5.72$\times$)   \\
\bottomrule
\end{tabular}
}

\vspace{-0.2cm}
\end{table}

It shows that our distilled student greatly accelerates the inference speed. Take ComplEx and RotatE as examples, the inference time of the 512-dimensional teachers on WN18RR is $7.03$ times and $7.81$ times of the 32-dimensional student. Compared with the teachers, the 64-dimensional students achieve \textbf{an average speed increase of 2.25$\times$, 2.22$\times$, 3.66$ \times$, and 3.98$ \times$}, and the 32-dimensional students achieve
\textbf{an average speed increase of 3.11$\times$, 3.35$\times$, 5.90$\times$, and 5.76$\times$} for TransE, SimplE, ComplEx and RotatE among the two datasets.

Previous experiments have proved that compared with the 512-dimensional teachers, our 64-dimensional students (8 times embedding CR) have little or no performance loss, and our 32-dimensional students (16 times embedding CR) retain most of performance. The results support that DualDE successfully reduces \textbf{7-15 times} embedding parameters and increase the inference speed by \textbf{2-6 times}.

\subsection{Q3: Do the soft label evaluation mechanism and two-stage distillation approach contribute and how much?} 
We conducted a series of ablation studies to evaluate the impact of the two proposed strategies of DualDE: the soft label evaluation mechanism and the two-stage distillation approach. 

First, to study the impact of the soft label evaluation mechanism,
we compare our method (DS) to that with removing the soft label evaluation mechanism (\emph{-SEM}). 
Then, to study the impact of the two-stage distillation approach,
we 
compare 
DS to that with removing the first stage (\emph{-S1}) and removing the second stage (\emph{-S2}).
Table~\ref{two-stage} summarizes the MRR and Hit@10 results on WN18RR. 

\begin{table}[htbp]
\caption{Ablation study on WN18RR. \textbf{D} refers to dimension of the student and \textbf{M} refers to method.}\label{two-stage}
\centering
\resizebox{0.48\textwidth}{!}{
\begin{tabular}{c | c|cc|cc|cc|cc}
\toprule
\multirow{2}{*}{\textbf{D}} &\multirow{2}{*}{ \textbf{M}} 
& \multicolumn{2}{c}{\textbf{TransE}}  
& \multicolumn{2}{c}{\textbf{SimplE}} 
& \multicolumn{2}{c}{\textbf{ComplEx}} 
& \multicolumn{2}{c}{\textbf{RotatE}} \\
&  & \emph{MRR} & \emph{H10} & \emph{MRR} & \emph{H10} & \emph{MRR} & \emph{H10} & \emph{MRR} & \emph{H10}  \\
\hline 
\multirow{4}{*}{64}
& DS &\textbf{.230} &\textbf{.528} &\textbf{.412} &\textbf{.485} &\textbf{.419} &\textbf{.507} &\textbf{.472} &\textbf{.568} \\
& \emph{-SEM} &.223 & .513 & .392 &.468 &.399 & .500 & .462 &.549\\
& \emph{-S1} &.227 & .526 & .407 &.481 &.417 & .506 & .463 &.564\\
& \emph{-S2} &.225 & .520 & .391 &.474 &.414 & .502 &.466 &.565\\
\hline
\multirow{3}{*}{32}
& DS &\textbf{.210} &\textbf{.484} &\textbf{.384} &\textbf{.479}  &\textbf{.397} &\textbf{.473} &\textbf{.468} &\textbf{.560} \\
& \emph{-SEM} &.189 & .449 & .352 &.464 &.362 & .451 & .447 &.524\\
& \emph{-S1} &.195 & .454 & .302 &.428 &.351 & .418 &.445 &.529\\
& \emph{-S2} &.202 & .466 & .357 &.455 &.385 & .464 &.461 &.558\\

\bottomrule
\end{tabular}
}

\end{table}

After removing \emph{SEM} (refer to -\emph{SEM}), all students' performance declines compared to DS. Among these four KGEs, the MRR and Hit@10 of 64-dimensional students drop by an average of $3.7\%$ and $2.8\%$, and the MRR and Hit@10 of 32-dimensional students drop by an average of $7.9\%$ and $5.4\%$. The results show that the soft label evaluation module, which evaluates the quality of the soft label for each triple and assigns different soft label and hard label weight to different triples, is indeed beneficial to the student model to master those difficult triples and get better performance. 

After removing \emph{S1} with only \emph{S2} preserved (refer to -\emph{S1}), the performance is overall lower than DS. Presumably, the reason is that both the teacher and the student will adapt to each other in \emph{S2}. With a randomly initialized student, the student conveys mostly useless information to the teacher which may be misleading and will crash the teacher. In addition, the performance of `\emph{-S1}' is very unstable. With `\emph{-S1}' setting, 64-dimensional students obtain results only slightly worse than DS, while 32-dimensional students  perform obviously very poor. For the 32-dimensional student of SimplE, the MRR and Hit@10 of `\emph{-S1}' drop by $21.4\%$ and $10.6\%$ compared with DS. This is even worse than using the most basic distillation method BKD, showing that the first stage is necessary for DualDE.

After removing \emph{S2} with only \emph{S1} preserved (refer to -\emph{S2}), the performance decreases in almost all setting. 
Compared with DS, the MRR of 64- and 32-dimensional students of `-\emph{S2}' decreased by an average of $2.4\%$ and $3.8\%$ among these four KGEs, indicating that the second stage can indeed make teacher and student adapt to each other, and further improve the result.

These results support the effectiveness of our two-stage distillation that first train the student in \emph{S1}  converging to a certain performance and then co-optimize the teacher and  student in \emph{S2}.

\section{Conclusion and Future Work}
Too many embedding parameters of the knowledge graph will bring huge storage and calculation challenges to actual application scenarios. In this work, we propose a novel KGE distillation method DualDE to compress KGEs into the lower-dimensional space to effectively transfer the knowledge of the teacher to the student. Considering the dual-influence between the teacher and the student, we propose two distillation strategies into DualDE: the soft label evaluation mechanism to adaptively assign different soft label and hard label weights to different triples and the two-stage distillation approach to enhance the student's acceptance of the teacher by encouraging them learn from each other. We have evaluated DualDE through link prediction task on several KGEs and benchmark datasets. Results show that our method can effectively reduce the embedding parameters and greatly improve the inference speed of a high-dimensional KGE with only a little or no performance loss.

In this work, we only consider KGE distillation from the perspective of a single modal, that is graph structure information of KG encoded by KGE methods. In the future, we would like to first explore the KGE distillation with multi-modal data, such as combining the graph structure information and the text (or image) information of the entity to further improve the performance of low-dimensional KGEs.


\begin{acks}
This work is funded by NSFC91846204/U19B2027, national key research program 2018YFB1402800.
\end{acks}

\newpage
\bibliographystyle{unsrt}
\bibliography{sample-sigconf.bbl}

\begin{thebibliography}{10}

\bibitem{6_DBLP:conf/emnlp/BerantCFL13}
Jonathan Berant, Andrew Chou, Roy Frostig, and Percy Liang.
\newblock Semantic parsing on freebase from question-answer pairs.
\newblock In {\em {EMNLP}}, pages 1533--1544. {ACL}, 2013.

\bibitem{7_DBLP:conf/acl/BerantL14}
Jonathan Berant and Percy Liang.
\newblock Semantic parsing via paraphrasing.
\newblock In {\em {ACL} {(1)}}, pages 1415--1425. The Association for Computer
  Linguistics, 2014.

\bibitem{8_DBLP:conf/acl/HoffmannZLZW11}
Raphael Hoffmann, Congle Zhang, Xiao Ling, Luke~S. Zettlemoyer, and Daniel~S.
  Weld.
\newblock Knowledge-based weak supervision for information extraction of
  overlapping relations.
\newblock In {\em {ACL}}, pages 541--550. The Association for Computer
  Linguistics, 2011.

\bibitem{9_DBLP:conf/i-semantics/DaiberJHM13}
Joachim Daiber, Max Jakob, Chris Hokamp, and Pablo~N. Mendes.
\newblock Improving efficiency and accuracy in multilingual entity extraction.
\newblock In {\em {I-SEMANTICS}}, pages 121--124. {ACM}, 2013.

\bibitem{10_DBLP:journals/corr/ZhangLHJLW016}
Yuanzhe Zhang, Kang Liu, Shizhu He, Guoliang Ji, Zhanyi Liu, Hua Wu, and Jun
  Zhao.
\newblock Question answering over knowledge base with neural attention
  combining global knowledge information.
\newblock {\em CoRR}, abs/1606.00979, 2016.

\bibitem{11_DBLP:conf/www/DiefenbachSM18}
Dennis Diefenbach, Kamal~Deep Singh, and Pierre Maret.
\newblock Wdaqua-core1: {A} question answering service for {RDF} knowledge
  bases.
\newblock In {\em {WWW} (Companion Volume)}, pages 1087--1091. {ACM}, 2018.

\bibitem{14_DBLP:conf/nips/BordesUGWY13}
Antoine Bordes, Nicolas Usunier, Alberto Garc{\'{\i}}a{-}Dur{\'{a}}n, Jason
  Weston, and Oksana Yakhnenko.
\newblock Translating embeddings for modeling multi-relational data.
\newblock In {\em {NIPS}}, pages 2787--2795, 2013.

\bibitem{12_DBLP:conf/icml/TrouillonWRGB16}
Th{\'{e}}o Trouillon, Johannes Welbl, Sebastian Riedel, {\'{E}}ric Gaussier,
  and Guillaume Bouchard.
\newblock Complex embeddings for simple link prediction.
\newblock In {\em {ICML}}, volume~48 of {\em {JMLR} Workshop and Conference
  Proceedings}, pages 2071--2080. JMLR.org, 2016.

\bibitem{19_DBLP:conf/iclr/SunDNT19}
Zhiqing Sun, Zhi{-}Hong Deng, Jian{-}Yun Nie, and Jian Tang.
\newblock Rotate: Knowledge graph embedding by relational rotation in complex
  space.
\newblock In {\em {ICLR} (Poster)}. OpenReview.net, 2019.

\bibitem{2_DBLP:journals/corr/HintonVD15}
Geoffrey~E. Hinton, Oriol Vinyals, and Jeffrey Dean.
\newblock Distilling the knowledge in a neural network.
\newblock {\em CoRR}, abs/1503.02531, 2015.

\bibitem{DBLP:conf/cvpr/ParkKLC19}
Wonpyo Park, Dongju Kim, Yan Lu, and Minsu Cho.
\newblock Relational knowledge distillation.
\newblock In {\em {CVPR}}, pages 3967--3976. Computer Vision Foundation /
  {IEEE}, 2019.

\bibitem{DBLP:conf/aaai/MirzadehFLLMG20}
Seyed{-}Iman Mirzadeh, Mehrdad Farajtabar, Ang Li, Nir Levine, Akihiro
  Matsukawa, and Hassan Ghasemzadeh.
\newblock Improved knowledge distillation via teacher assistant.
\newblock In {\em {AAAI}}, pages 5191--5198. {AAAI} Press, 2020.

\bibitem{DBLP:conf/www/Wang0MS21}
Kai Wang, Yu~Liu, Qian Ma, and Quan~Z. Sheng.
\newblock Mulde: Multi-teacher knowledge distillation for low-dimensional
  knowledge graph embeddings.
\newblock In {\em {WWW}}, pages 1716--1726. {ACM} / {IW3C2}, 2021.

\bibitem{5_DBLP:conf/emnlp/SunCGL19}
Siqi Sun, Yu~Cheng, Zhe Gan, and Jingjing Liu.
\newblock Patient knowledge distillation for {BERT} model compression.
\newblock In {\em {EMNLP/IJCNLP} {(1)}}, pages 4322--4331. Association for
  Computational Linguistics, 2019.

\bibitem{18_DBLP:conf/icml/NickelTK11}
Maximilian Nickel, Volker Tresp, and Hans{-}Peter Kriegel.
\newblock A three-way model for collective learning on multi-relational data.
\newblock In {\em {ICML}}, pages 809--816. Omnipress, 2011.

\bibitem{22_DBLP:journals/corr/YangYHGD14a}
Bishan Yang, Wen{-}tau Yih, Xiaodong He, Jianfeng Gao, and Li~Deng.
\newblock Embedding entities and relations for learning and inference in
  knowledge bases.
\newblock In {\em {ICLR} (Poster)}, 2015.

\bibitem{DBLP:conf/nips/Kazemi018}
Seyed~Mehran Kazemi and David Poole.
\newblock Simple embedding for link prediction in knowledge graphs.
\newblock In {\em NeurIPS}, pages 4289--4300, 2018.

\bibitem{25_DBLP:conf/aaai/WangZFC14}
Zhen Wang, Jianwen Zhang, Jianlin Feng, and Zheng Chen.
\newblock Knowledge graph embedding by translating on hyperplanes.
\newblock In {\em {AAAI}}, pages 1112--1119. {AAAI} Press, 2014.

\bibitem{26_DBLP:conf/aaai/LinLSLZ15}
Yankai Lin, Zhiyuan Liu, Maosong Sun, Yang Liu, and Xuan Zhu.
\newblock Learning entity and relation embeddings for knowledge graph
  completion.
\newblock In {\em {AAAI}}, pages 2181--2187. {AAAI} Press, 2015.

\bibitem{27_DBLP:conf/acl/JiHXL015}
Guoliang Ji, Shizhu He, Liheng Xu, Kang Liu, and Jun Zhao.
\newblock Knowledge graph embedding via dynamic mapping matrix.
\newblock In {\em {ACL} {(1)}}, pages 687--696. The Association for Computer
  Linguistics, 2015.

\bibitem{20_DBLP:conf/nips/0007TYL19}
Shuai Zhang, Yi~Tay, Lina Yao, and Qi~Liu.
\newblock Quaternion knowledge graph embeddings.
\newblock In {\em NeurIPS}, pages 2731--2741, 2019.

\bibitem{21_DBLP:conf/acl/XuL19}
Canran Xu and Ruijiang Li.
\newblock Relation embedding with dihedral group in knowledge graph.
\newblock In {\em {ACL} {(1)}}, pages 263--272. Association for Computational
  Linguistics, 2019.

\bibitem{35_DBLP:conf/acl/Sachan20}
Mrinmaya Sachan.
\newblock Knowledge graph embedding compression.
\newblock In {\em {ACL}}, pages 2681--2691. Association for Computational
  Linguistics, 2020.

\bibitem{DBLP:conf/iccv/GongLJLHLYY19}
Ruihao Gong, Xianglong Liu, Shenghu Jiang, Tianxiang Li, Peng Hu, Jiazhen Lin,
  Fengwei Yu, and Junjie Yan.
\newblock Differentiable soft quantization: Bridging full-precision and low-bit
  neural networks.
\newblock In {\em {ICCV}}, pages 4851--4860. {IEEE}, 2019.

\bibitem{32_DBLP:journals/tnn/CastellanoFP97}
Giovanna Castellano, Anna~Maria Fanelli, and Marcello Pelillo.
\newblock An iterative pruning algorithm for feedforward neural networks.
\newblock {\em {IEEE} Trans. Neural Networks}, 8(3):519--531, 1997.

\bibitem{33_DBLP:conf/iclr/MolchanovTKAK17}
Pavlo Molchanov, Stephen Tyree, Tero Karras, Timo Aila, and Jan Kautz.
\newblock Pruning convolutional neural networks for resource efficient
  inference.
\newblock In {\em {ICLR} (Poster)}. OpenReview.net, 2017.

\bibitem{34_DBLP:conf/icml/LinTA16}
Darryl~Dexu Lin, Sachin~S. Talathi, and V.~Sreekanth Annapureddy.
\newblock Fixed point quantization of deep convolutional networks.
\newblock In {\em {ICML}}, volume~48 of {\em {JMLR} Workshop and Conference
  Proceedings}, pages 2849--2858. JMLR.org, 2016.

\bibitem{36_DBLP:conf/iclr/DehghaniGVUK19}
Mostafa Dehghani, Stephan Gouws, Oriol Vinyals, Jakob Uszkoreit, and Lukasz
  Kaiser.
\newblock Universal transformers.
\newblock In {\em {ICLR} (Poster)}. OpenReview.net, 2019.

\bibitem{37_DBLP:conf/iclr/LanCGGSS20}
Zhenzhong Lan, Mingda Chen, Sebastian Goodman, Kevin Gimpel, Piyush Sharma, and
  Radu Soricut.
\newblock {ALBERT:} {A} lite {BERT} for self-supervised learning of language
  representations.
\newblock In {\em {ICLR}}. OpenReview.net, 2020.

\bibitem{39_DBLP:journals/corr/abs-1903-12136}
Raphael Tang, Yao Lu, Linqing Liu, Lili Mou, Olga Vechtomova, and Jimmy Lin.
\newblock Distilling task-specific knowledge from {BERT} into simple neural
  networks.
\newblock {\em CoRR}, abs/1903.12136, 2019.

\bibitem{44_DBLP:conf/naacl/DevlinCLT19}
Jacob Devlin, Ming{-}Wei Chang, Kenton Lee, and Kristina Toutanova.
\newblock {BERT:} pre-training of deep bidirectional transformers for language
  understanding.
\newblock In {\em {NAACL-HLT} {(1)}}, pages 4171--4186. Association for
  Computational Linguistics, 2019.

\bibitem{38_DBLP:conf/iclr/TianKI20}
Yonglong Tian, Dilip Krishnan, and Phillip Isola.
\newblock Contrastive representation distillation.
\newblock In {\em 8th International Conference on Learning Representations,
  {ICLR} 2020, Addis Ababa, Ethiopia, April 26-30, 2020}. OpenReview.net, 2020.

\bibitem{45_DBLP:journals/corr/abs-1909-11687}
Sanqiang Zhao, Raghav Gupta, Yang Song, and Denny Zhou.
\newblock Extreme language model compression with optimal subwords and shared
  projections.
\newblock {\em CoRR}, abs/1909.11687, 2019.

\bibitem{DBLP:journals/jmlr/AchilleS18}
Alessandro Achille and Stefano Soatto.
\newblock Emergence of invariance and disentanglement in deep representations.
\newblock {\em J. Mach. Learn. Res.}, 19:50:1--50:34, 2018.

\bibitem{16_DBLP:conf/emnlp/ToutanovaCPPCG15}
Kristina Toutanova, Danqi Chen, Patrick Pantel, Hoifung Poon, Pallavi
  Choudhury, and Michael Gamon.
\newblock Representing text for joint embedding of text and knowledge bases.
\newblock In {\em {EMNLP}}, pages 1499--1509. The Association for Computational
  Linguistics, 2015.

\bibitem{15_DBLP:conf/aaai/DettmersMS018}
Tim Dettmers, Pasquale Minervini, Pontus Stenetorp, and Sebastian Riedel.
\newblock Convolutional 2d knowledge graph embeddings.
\newblock In {\em {AAAI}}, pages 1811--1818. {AAAI} Press, 2018.

\bibitem{DBLP:conf/icassp/WuCW19}
Meng{-}Chieh Wu, Ching{-}Te Chiu, and Kun{-}Hsuan Wu.
\newblock Multi-teacher knowledge distillation for compressed video action
  recognition on deep neural networks.
\newblock In {\em {ICASSP}}, pages 2202--2206. {IEEE}, 2019.

\bibitem{DBLP:conf/emnlp/HanCLLLSL18}
Xu~Han, Shulin Cao, Xin Lv, Yankai Lin, Zhiyuan Liu, Maosong Sun, and Juanzi
  Li.
\newblock Openke: An open toolkit for knowledge embedding.
\newblock In {\em {EMNLP} (Demonstration)}, pages 139--144. Association for
  Computational Linguistics, 2018.

\bibitem{49_DBLP:journals/corr/KingmaB14}
Diederik~P. Kingma and Jimmy Ba.
\newblock Adam: {A} method for stochastic optimization.
\newblock In {\em {ICLR} (Poster)}, 2015.

\end{thebibliography}


\end{document}